\documentclass[10pt,twocolumn,letterpaper]{article}

\usepackage{iccv}
\usepackage{times}
\usepackage{epsfig}
\usepackage{graphicx}
\usepackage{amsmath}
\usepackage{amssymb}

\newtheorem{theorem}{Theorem}
\usepackage{here}

\usepackage{mydefe}

\def\bm#1{\mbox{\boldmath $#1$}}
\def\vecg#1{{\bm #1}}

\def\vec#1{{\rm\bf #1}}
\def\mtr#1{{\rm\bf #1}}

\def\defas{:=}

\def\vectorize{\mathop{\textrm{vec}}\nolimits}

\usepackage{algorithm}
\usepackage{algorithmic}
\renewcommand{\algorithmicrequire}{\textbf{Input:}}
\renewcommand{\algorithmicensure}{\textbf{Output:}}
\algsetup{linenosize=\normalsize,linenodelimiter=\ }
\renewcommand{\algorithmiccomment}[1]{// #1}

\iccvfinalcopy 

\def\iccvPaperID{}

\ificcvfinal\fi
\begin{document}

\title{Multiple Pattern Classification by Sparse Subspace Decomposition}

\author{Tomoya Sakai\\
Institute of Media and Information Technology, Chiba University\\
1-33 Yayoi, Inage, Chiba, Japan\\
{\tt\small tsakai@faculty.chiba-u.jp}
}

\maketitle

\begin{abstract}
A robust classification method is developed
on the basis of sparse subspace decomposition.
This method tries to decompose
a mixture of subspaces of unlabeled data (queries)
into class subspaces as few as possible.
Each query is classified into the class 
whose subspace significantly contributes to the decomposed subspace.
Multiple queries from different classes
can be simultaneously classified into their respective classes.
A practical greedy algorithm of the sparse subspace decomposition
is designed 
for the classification.
The present method achieves high recognition rate and robust performance
exploiting 
joint sparsity.
\end{abstract}

\section{Introduction}

Classification is a task of assigning
one or more class labels to unlabeled data (query data).
A collection of labeled data (training data)
is available for the classification.
The patterns or signals to be classified are usually
groups of measurement data expressed as high-dimensional vectors.

Depending on purposes,
we need pattern classifiers that can answer
\begin{itemize}
 \item a label to each of queries, 
 \item a label to a set of queries, 
 \item a few labels to each of queries, 
 \item a label ``invalid'' to an unclassifiable query.
\end{itemize}
We develop a framework of using subspaces
for all these functionalities. 
We regard the unlabeled data as a mixture of subspaces.
The key idea is to decompose it into the subspaces of classes
as few as possible.
Only the classes explaining concisely the mixture
are relevant to the unlabeled data.
In the classification,
the unlabeled data are usually supposed to belong
to a few (typically one) classes.
Therefore, the classification process can be
interpreted as sparse decomposition of the subspace mixture.

This work is inspired by the recently developing field of compressed sensing
\cite{Donoho06,Candes06a,Candes06b,Candes08RIP,Candes08intro}
and its innovative applications
to robust face recognition \cite{Wright08},
action recognition \cite{Yang09},
computer vision and image processing \cite{Wright09}.
The essential idea of these works is to exploit the prior knowledge
that a signal is sparse and compressible.
The theory of compressed sensing
is very helpful and informative for us to answer questions such as
``How many measurements are enough for the pattern recognition?'' and
``What is the role of feature extraction?''
It is worthy to explore the potential of sparse decomposition
for substantial improvement of the subspace methods.

The rest of this paper is organized as follows.
Section \ref{sec:preliminaries} provides
preliminary details and definitions of subspace representation
for sparse decomposition.
In Section \ref{sec:classification},
we propose a classification method named {\it sparse subspace method},
which exploits the sparseness property for the classification tasks
described above.
A practical algorithm of the sparse subspace decomposition
is presented in Section \ref{sec:decomposition}.
We show some tentative evaluation results of
the sparse subspace method using a face database
in Section \ref{sec:experiments} before concluding in Section \ref{sec:conclusion}.

\section{Preliminaries}
\label{sec:preliminaries}

Let $\mtr S_k\in\mathbb{R}^{d\times n_k}$
be a matrix of training dataset of $k$-th class ($k=1,\dots,C$),
in which $n_k$ labeled patterns are represented as
the $d$-dimensional column feature vectors.
We describe as follows the linear subspaces, their union,
block sparsity, and sparse linear representation of a subspace.
We also define a classification space
where the sparsity should be encouraged.

\paragraph{Linear subspaces of training datasets}

The class subspace is defined as
a vector subspace whose elements are the feature vectors
of labeled data.
We describe the subspace as a vector subspace
in the normed space:
\begin{equation}
\mathcal{S}_k\defas\span\mtr S_k\subset(\mathbb{R}^d,l^2).
\end{equation}
$S_k$ approximates the $k$-th class subspace.
We denote the dimensionality of $\mathcal{S}_k$
by $\dim\mathcal{S}_k=\rank\mtr S_k$.

\paragraph{Union of subspaces}

The union of subspaces is the subspace obtained by
combining the feature vectors of each class.
\begin{equation}
\mathcal{S}\defas\cup_{k=1}^C\mathcal{S}_k=\span\mtr S
\subseteq(\mathbb{R}^d,l^2)
\end{equation}
Here, $\mtr S$ is the concatenation of $\mtr S_k$ as
\begin{equation}
 \mtr S\defas[\mtr S_1,\dots,\mtr S_C]\in\mathbb{R}^{d\times N}
\label{eq:concatenated training datasets}
\end{equation}
and $N\defas\sum_{k=1}^C n_k$.
The dimensionality of $\mathcal{S}$ is denoted by
$\dim\mathcal{S}=\rank\mtr S$.

We say that the subspaces $\mathcal{S}_k$ ($k=1,\dots,C$)
are independent
if and only if
any subspace $\mathcal{S}_k$ is not a subset of
the union of the other subspaces, \ie,
$\mathcal{S}_k\not\subset\cup_{i\neq k}^C\mathcal{S}_i$
for $\forall k$.

\paragraph{Linear representation of vector(s)}

Given sufficient training dataset,
a $d$-dimensional vector $\vec q$ of unlabeled data
(hereafter ``query'' vector)
will be approximately represented
as a linear combination of vectors from class subspaces.
\begin{equation}
\vec q
=\sum_{k=1}^C\mtr S_k\vecg\alpha_k
=\mtr S\vecg\alpha
\label{eq:linear representation}
\end{equation}
Here, $\vecg\alpha_k\in(\mathbb{R}^{n_k},l^2)$
is a vector of coefficients corresponding to the $k$-th class,
and
\begin{equation}
\vecg\alpha\defas\bmatrix{c}{\vecg\alpha_1\\ \vdots\\ \vecg\alpha_C}\in(\mathbb{R}^N,l^2)
\label{eq:concatenated coefficients}
\end{equation}
is the concatenation of $\vecg\alpha_k$.

If a set of queries is given as a matrix
\begin{equation}
\mtr Q\defas[\vec q^{(1)},\dots,\vec q^{(n)}]\in\mathbb{R}^{d\times n},
\label{eq:query matrix}
\end{equation}
then we will solve
\begin{equation}
 \mtr Q=\mtr S\mtr A.
\label{eq:linear representation of vectors}
\end{equation}
Here,
\begin{equation}
 \mtr A\defas[\vecg\alpha^{(1)},\dots,\vecg\alpha^{(n)}]
\in\mathbb{R}^{N\times n}
\end{equation}
is the matrix of unknown coefficients, and
\begin{equation}
\vecg\alpha^{(j)}\defas\bmatrix{c}{\vecg\alpha_1^{(j)}\\ \vdots\\ \vecg\alpha_C^{(j)}}\in\mathbb{R}^N
\end{equation}
is the concatenated vector of coefficients for the $j$-th query.
The matrix $\mtr A$ can also be described as
\begin{equation}
 \mtr A=\bmatrix{c}{\mtr A_1\\ \vdots\\ \mtr A_C}
\label{eq:row stacked form}
\end{equation}
where
\begin{equation}
 \mtr A_k\defas [\vecg\alpha_k^{(1)},\dots,\vecg\alpha_k^{(n)}]
\in\mathbb{R}^{n_k\times n}.
\end{equation}

The systems of linear equations
as (\ref{eq:linear representation of vectors})
is called the problem for multiple measurement vectors (MMV),
while the case of a single measurement $n=1$
as (\ref{eq:linear representation})
is referred to as SMV
\cite{Chen05,Cotter05,Eldar08}.
The query vectors correspond to the measurements in this context.

\paragraph{Uniqueness}

The solution $\vecg\alpha$ to (\ref{eq:linear representation})
or $\mtr A$ to (\ref{eq:linear representation of vectors})
exists if and only if 
\begin{equation}
\vec q^{(j)}\in\mathcal{S}
\;\;\forall j,
\label{eq:existence condition}
\end{equation}
\ie, the queries lie on the union of class subspaces.
For $\dim\mathcal{S}<d$,
the solution does not always exist.
The solution may be dense even if it exists.
Most components are nonzero despite the fact
that at most $n$ class subspaces are relevant to $n$ queries.
This problem is due to invalid situation
where training datasets are insufficient
to identify the class, uniquely.

The actual problem we should cope with
is the underdetermined case $d=\dim\mathcal{S}<N$, \ie,
the dimensionality of the union of subspaces
is less than the total number $N$
of training samples.
Unless the training data matrices $\mtr S_k$ are rank-degenerated
so that $\dim\mathcal{S}<d$,
the $C$ subspaces of training data cannot be independent
in the $d$-dimensional space.
There is an infinite number of ways
to express the query vector 
by the linear combination of the subspace bases.
The underdetermined problem requires regularization
to select a unique solution.
A sparse solution indicating relevant classes
would be preferable.

\paragraph{Block sparsity}

A vector $\vecg\xi\in(\mathbb{R}^N,l^0)$ is called $m$-sparse
if $||\vecg\xi||_0\leq m$.
Here, $||\cdot||_0$ denotes the $l^0$ norm,
which counts the nonzero vector components.
As the support of a function is
the subset of its domain where it is nonzero,
the support of a vector $\vecg\xi$
is defined as $\mathcal{T}=\{i|\xi_i\neq 0\}$.
The $l^0$ norm is the cardinality of the support.

We define a block-wise sparsity level in a similar manner
to \cite{Eldar08}.
Let $f_{\mathcal{N}}$ be a map
from $\forall\mtr X\in(\mathbb{R}^{N\times n},l^F)$
to $\vecg\gamma\in(\mathbb{R}_{+}^C,l^0)$
according to a list $\mathcal{N}\defas\{n_1,\dots,n_C\}$ such that
\begin{equation}
f_{\mathcal{N}} : 
\bmatrix{c}{\mtr X_1\\ \vdots\\ \mtr X_C}
\rightarrow
\bmatrix{c}{ ||\mtr X_1||_F\\ \vdots \\ ||\mtr X_C||_F }\defas\vecg\gamma.
\label{eq:map to l0 classification space}
\end{equation}
Here, $\mtr X_k\in(\mathbb{R}^{n_k\times n},l^F)$
is the $k$-th row block of $\mtr X$ with respect to $\mathcal{N}$, and
$||\cdot||_F$ denotes the Frobenius norm $l^F$.
Clearly, 
\begin{equation}
f_{\mathcal{N}} : 
\bmatrix{c}{\vec x_1\\ \vdots\\ \vec x_C}
\rightarrow
\bmatrix{c}{ ||\vec x_1||_2\\ \vdots \\ ||\vec x_C||_2 }
\end{equation}
for $n=1$.
A vector $\vec x\in(\mathbb{R}^N,l^2)$ is called block $M$-sparse
over $\mathcal{N}$
if $\vec x_k\neq \vec 0$ for at most
$M$ indices $k$.
The block sparsity is measured as
\begin{equation}
 ||\vec x||_{0,\mathcal{N}}
\defas
||f_{\mathcal{N}}(\vec x)||_0.
\label{eq:block sparsity}
\end{equation}
That is, $||\cdot||_{0,\mathcal{N}}$ counts the number of nonzero blocks.
We measure the row block sparsity of a matrix
$\mtr X\in(\mathbb{R}^{N\times n},l^F)$
over $\mathcal{N}$ as
\begin{equation}
||\mtr X||_{0,\mathcal{N}}
\defas
||f_{\mathcal{N}}(\mtr X)||_0.
\label{eq:row block sparsity}
\end{equation}
A matrix $\mtr X$ is row block $M$-sparse
if $||\mtr X||_{0,\mathcal{N}}\leq M$.

We remark that the row block $M$-sparse matrix $\mtr X$
can be converted into a block $M$-sparse vector $\vectorize(\mtr X^\top)$.
Here, the operator $\vectorize$ transforms a matrix into
a column vector by stacking all the columns of the matrix.
For $\mathcal{N}\defas\{n_1,\dots,n_C\}$
and $\mathcal{N}'\defas\{nn_1,\dots,nn_C\}$,
the block sparsity of $\mtr X\in\mathbb{R}^{N\times n}$
over $\mathcal{N}$ is preserved as
\begin{equation}
 ||\mtr X||_{0,\mathcal{N}}=||\vectorize(\mtr X^\top)||_{0,\mathcal{N}'}.
\end{equation}

\paragraph{Sparse representation of subspace}

In the underdetermined case,
the columns of matrix $\mtr S\in\mathbb{R}^{d\times N}$
represent an overcomplete basis of $\mathbb{R}^d$ for $d<N$.
Equation (\ref{eq:linear representation})
and (\ref{eq:linear representation of vectors})
can be consistent with infinitely many solutions
$\vecg\alpha$ and $\mtr A$, respectively.

We denote the subspace of query vector(s)
by $\mathcal{Q}=\span\vec q$ or $\span\mtr Q$.
If a possible solution $\vecg\alpha$ or $\mtr A$ is block sparse
over $\mathcal{N}=\{n_1,\dots,n_C\}$,
the query subspace $\mathcal{Q}$ consists
of a small minority of class subspaces
corresponding to nonzero $\vecg\alpha_k$ or $\mtr A_k$.
In other words, the query subspace is sparsely represented
by the class subspaces.
The sparsity of the subspace representation
can be quantified as $||\vecg\alpha||_{0,\mathcal{N}}$
or $||\mtr A||_{0,\mathcal{N}}$.

\paragraph{Classification space}

By definition,
the block sparsity $||\vecg\alpha||_{0,\mathcal{N}}$
or $||\mtr A||_{0,\mathcal{N}}$
is measured by the $l^0$ norm
of the $C$-dimensional vector
$\vecg\gamma\defas f_{\mathcal{N}}(\vecg\alpha)$
or $f_{\mathcal{N}}(\mtr A)$.
The components of $\vecg\gamma$ imply
the degrees of class membership.
The sparser $\vecg\gamma$ is,
the more certainly the class label of each query is identified.
The sparsity is properly measured by the $l^0$ norm.
Therefore,
we refer to the normed space $\mathcal{C}=(\mathbb{R}_{+}^C,l^0)$,
where $\vecg\gamma$ resides, as the classification space.

\section{Classification based on sparse subspace representation}
\label{sec:classification}

From the viewpoint of classification,
each query vector is supposed to be composed only of
vectors from the subspace of a class to which the query is classified.
The subspace spanned by the query vectors
should be 
represented
as sparsely as possible by the class subspaces concerned with the queries.
In our notation,
the $C$-dimensional vector in the classification space,
$\vecg\gamma\defas f_{\mathcal{N}}(\vecg\alpha)$ or $f_{\mathcal{N}}(\mtr A)$,
is intended to be sparsest.
The sparsity is properly measured by the $l^0$ norm of $\vecg\gamma$.
Therefore, we incorporate 
minimization of the $l^0$ norm in the classification framework.

\subsection{Formulation}
\label{subsec:formulation}

Let $\mtr S\in\mathbb{R}^{d\times N}$
be the concatenation of
$\mtr S_k\in\mathbb{R}^{d\times n_k}$
($k=1,\dots,C$, $d=\rank\mtr S<N=\sum_{k=1}^Cn_k$), \ie,
the matrices of training datasets.
Given the matrix $\mtr Q\in\mathbb{R}^{d\times n}$ of $n$ query vectors,
we solve the $l^0$-minimization problem:
\begin{equation}
 \min_{\mtr A} ||\mtr A||_{0,\mathcal{N}}
\quad\mbox{subject to}\quad \mtr Q=\mtr S\mtr A.
\label{eq:sparse decomposition by minimization}
\end{equation}
Here, $\mathcal{N}$ specifies the sizes of row blocks for sparsification.
Typically, $\mathcal{N}=\{n_1,\dots,n_C\}$.
The matrix $\mtr A$ is released from being row-block sparse
if $\mathcal{N}=\mathcal{N}_1\defas\{\forall n_i\!=\!1,i\!=\!1,\dots,N\}=\{1,\dots,1\}$.

One can rewrite
the problem (\ref{eq:sparse decomposition by minimization}) as
\begin{eqnarray}
&& \min_{\mtr A} ||\vectorize(\mtr A^\top)||_{0,\mathcal{N}'}
\quad\mbox{subject to}\quad\nonumber\\
&&\qquad\qquad\vectorize(\mtr Q^\top)=(\mtr S\otimes\mtr I_n)\vectorize(\mtr A^\top)
\label{eq:sparse decomposition by minimization vectorized}
\end{eqnarray}
where $\otimes$ denotes the Kronecker product, and
$\mtr I_n$ is the identity matrix of size $n$.
The list $\mathcal{N}'$
defines the block sizes of
the $nN$-dimensional vector $\vectorize(\mtr A^\top)$.

The $l^0$-minimization problem
(\ref{eq:sparse decomposition by minimization vectorized})
is well investigated in the literature \cite{Eldar08}.
The uniqueness of the solution is
guaranteed under the condition
called block restricted isometry property (block RIP).
Assuming $\vec q^{(j)}\in\mathcal{S}$,
the RIP condition for our problem can be described as
\begin{eqnarray}
&& (1-\delta_{M|\mathcal{N}'})||\vec v||_2^2
\nonumber\\
&&\qquad\leq
||(\mtr S\otimes\mtr I_n)\,\vec v||_2^2
\nonumber\\
&&\qquad\qquad\leq
(1+\delta_{M|\mathcal{N}'})||\vec v||_2^2\quad
\forall\vec v\in{\mathbb{R}^{nN}}.
\label{eq:block RIP condition}
\end{eqnarray}
where $\delta_{M|\mathcal{N}'}$
is called the block-RIP constant
dependent on the block sparsity $M$ over $\mathcal{N}'$.
In practice, we normalize the blocks $\mtr S_k$
in order for the matrix $\mtr S\otimes\mtr I_n$
to satisfy the condition.
The block RIP condition is less stringent
than the standard RIP condition, which is widely used
in the field of compressed sensing
\cite{Donoho06,Candes06a,Candes06b,Candes08RIP,Candes08intro}.

\subsection{Dimensionality reduction}

In (\ref{eq:sparse decomposition by minimization}),
we assume the linear system $\mtr Q=\mtr S\mtr A$
to be underdetermined as $d=\rank\mtr S<N$,
and regularize it by the $l^0$ minimization.
Actually, we do not have to deal with the queries and training data
in a space of dimension $d\geq N$.
The recent works in the emerging area of compressed sensing
show that a small number of projections
of a sparse vector 
can contain its salient information
enough to recover the vector 
with regularization
that promotes sparsity \cite{Donoho06,Candes06b,Candes06c}.
The statements in \cite{Candes05,Rudelson05}
guaranteeing the recovery are described as follows.
\begin{theorem}
Let $\vec x\defas\mtr\Psi^\top\vec s$ be a $d$-dimensional vector
represented by a $m$-sparse vector $\vec s\in\mathbb{R}^d$
using a basis $\mtr\Psi^\top\in\mathbb{R}^{d\times d}$.
Then, $\vec s$ can be reconstructed
from a $\hat d$-dimensional vector
$\hat{\vec x}\defas\mtr\Phi\vec x$
with probability $1-e^{-\mathcal{O}(\hat d)}$.
Here, $\mtr\Phi\in\mathbb{R}^{\hat d\times d}$
is a random matrix and
$\hat d\geq\hat d_0\defas\mathcal{O}(m\log(d/m))$.
\label{th:measurement for reconstruction}
\end{theorem}
Specially, $\hat d\geq 2m\log(d/\hat d)$
holds if $m\ll d$ \cite{Candes06,Donoho09}.
It is also possible to recover the sparse vector $\vec s$ from
a small number of projections, $\hat{\vec x}$,
with overwhelming probability
in more general case
where $\mtr\Phi$ and $\mtr\Psi$ are incoherent
\cite{Candes06,Candes07,Candes08intro}.

The reconstructability
in Theorem \ref{th:measurement for reconstruction}
suggests that one can obtain
the $d$-dimensional
$m$-sparse solution 
from a much lower $\hat d$-dimensional vector
after linear transformation.
Wright \etal \cite{Wright08} showed,
in their framework of face recognition
based on sparse representation,
that the computational cost is reduced
without significant loss of recognition rate
by linear transformations
into lower dimensional feature spaces,
such as Eigenfaces, Fisherfaces, Laplacianfaces,
downsampling, and random projection.
These transformations act as dimensionality reduction that
preserves information for the recognition.
Especially, random projection is
a data-independent dimensionality reduction technique,
and one can exactly recover the original $d$-dimensional vector.
For this reason,
we employ the dimensionality reduction
if $d$ is too high for computation.

\subsection{Classifiers}

\paragraph{$n$-to-one classifier}

Since the minimizer $\mtr A$ for (\ref{eq:sparse decomposition by minimization})
is a row block $M$-sparse matrix,
the $M$ blocks indicate
the $M_C$ $(M_C\leq M$) classes concerned with the query subspaces.
For the task of classifying all $n$ queries into one class ($M_C=1$),
we calculate the residuals $r_k$
of the representations by the class subspaces.
\begin{equation}
r_k(\mtr Q;\mtr A)\defas||\mtr Q-\mtr S_k\mtr A_k||_F.
\label{eq:residual}
\end{equation}
The residuals quantify the dissimilarities between
the query subspace and the class subspaces.
Note that most of the residuals are $||\mtr Q||_F$
because of the sparsity.
If the query subspace $\mathcal{Q}$
can be approximately represented
by one of the class subspaces,
the class label is identified as
\begin{equation}
 \arg\min_k r_k(\mtr Q;\mtr A).
\label{eq:n-to-one classifier}
\end{equation}
This classification method achieves the same task as
the mutual subspace methods \cite{Maeda85,Yamaguchi98,Fukui03}
in a fundamentally different strategy.
The mutual subspace methods are robust
owing to the multiple queries.
The robustness is further enhanced by the block sparsification
in our scheme.
The $l^0$ minimization
in (\ref{eq:sparse decomposition by minimization})
encourages the vector of class membership degrees,
$f_{\mathcal{N}}(\mtr A)$,
to be as sparse as possible in the classification space.
For the underdetermined problem with a sparse solution,
the recent works in the emerging area of compressed sensing
\cite{Donoho06,Candes06a,Candes06b,Candes08RIP}
prove the exact recovery under the $l^0$ or $l^1$ regularization.
Since the $l^0$ / $l^1$ minimizer is very insensitive to outliers,
the sparse representation is robust compared to
the conventional representations by $l^2$-based regularization
\eg PCA.

We also remark that 
if $n=1$ and $\mathcal{N}=\mathcal{N}_1$,
our $n$-to-one classification
is exactly the same as the sparse representation-based
classification (SRC) proposed in \cite{Wright08}.
Our classification based on sparse subspace representation is
therefore an extension of the SRC for multiple queries.

\paragraph{$n$-to-ones classifier}

It is also possible to classify $n$ queries
into their respective classes.
We calculate $C\times n$ residual matrix
whose $kj$-th entry measures the dissimilarity between
the $j$-th query and its reconstruction in the $k$-th subspace:
\begin{equation}
r_k^{(j)}(\mtr Q;\mtr A)\defas
||\vec q^{(j)}-\mtr S_k\vecg\alpha_k^{(j)}||_2.
\label{eq:residual vector}
\end{equation}
Note that most of the residual entries are $||\vec q^{(j)}||_2$
because of the sparsity.
If the query subspace $\mathcal{Q}$ can be approximately
represented by union of a small number of class subspaces,
the class label for the $j$-th query is identified as
\begin{equation}
 \arg\min_k r_k^{(j)}(\mtr Q;\mtr A).
\label{eq:multi-classifier}
\end{equation}
Again, our method is expected to be robust owing to the multiple queries.
Furthermore,
the classes irrelevant to the queries are strongly excluded
by the $l^0$ minimization.
Therefore, the classifier (\ref{eq:multi-classifier})
can detect the respective class for each query
without giving the number of relevant classes.

\paragraph{$n$-to-$M$ classifier}

Let us mention the potential of the sparse subspace representation
for finding $n$-to-$M$ relations,
although we do not go into the detail of this type of multiple classification
in this paper.
If a query simultaneously belongs to multiple classes,
the query vector is represented as
a linear combination of vectors from the subspaces of the relevant classes.
The residuals $r_k^{(j)}$ for such query cannot be zero,
but the relevant classes are found by thresholding $r_k^{(j)}$.
Thus, each of $n$ queries is assigned to some of $M$ classes.

\paragraph{Classification validity}

A classifier should answer ``invalid''
if the given query belongs to an unknown class.
As suggested in \cite{Wright08},
such an unclassifiable query is perceived to be so
by measuring how the nonzero components of $\mtr A$
concentrate on a single class.
Wright \etal defined the sparsity concentration index (SCI),
which quantifies the validity of the classification \cite{Wright08}.
One may compute the SCI for each column of $\mtr A$
to validate the corresponding query.

\subsection{Sparse subspace method}

Our classification method based on the sparse subspace representation
is summarized in Algorithm \ref{alg:SSM}.

\begin{algorithm}[htb]
\caption{Sparse subspace method (SSM)}
\label{alg:SSM}
\begin{algorithmic}[1]
\REQUIRE
$\mtr Q\in\mathbb{R}^{d\times n}$: matrix of $n$ queries
as (\ref{eq:query matrix}),\quad
$\mtr S\in\mathbb{R}^{d\times N}$: concatenated matrix of training datasets as (\ref{eq:concatenated training datasets}),\quad
$\mathcal{N}$: list of row block sizes;
\ENSURE
$\mathcal{L}$: set of class labels;
\STATE
perform dimensionality reduction of $\mtr Q$
and $\mtr S$ if $d$ is intractably high;\label{step:dimensionality reduction}
\STATE
normalize the columns of $\mtr S$ to have unit $l^2$ norm;
\STATE
decompose $\mtr Q$ with respect to $\mtr S$
to obtain the sparse subspace representation.
\label{step:sparse decomposition}
\STATE
find the class label $\mathcal{L}=\{\arg\min_k r_k(\mtr Q;\mtr A) \}$ or\\
$\mathcal{L}=\{\arg\min_k r_k^{(1)}(\mtr Q;\mtr A),\dots,\arg\min_k r_k^{(n)}(\mtr Q;\mtr A)\}$.
\end{algorithmic}
\end{algorithm}

The major concern is the sparse subspace decomposition
of $\mathcal{Q}$ at Step \ref{step:sparse decomposition}.
In the next section,
we present a practical algorithm of the decomposition, SSD-ROMP,
which efficiently and stably provides
approximate solution to (\ref{eq:sparse decomposition by minimization}).

\section{Sparse subspace decomposition}
\label{sec:decomposition}

The sparse decomposition of $\mtr Q$
in (\ref{eq:sparse decomposition by minimization})
is considered as a MMV problem
whose solution is row-block sparse.
The solution has two important characteristics:
the column vectors $\vecg\alpha^{(j)}$ of $\mtr A$
share nonzero blocks as their support,
and the block partitions are fixed by $\mathcal{N}$ in advance.

\subsection{Prior work on MMV}

Configuration of the nonzero entries in the solution $\mtr A$
is called the joint sparsity model (JSM) \cite{Baron05,Marco05a}.
There are some prior works on the MMV problems with
several JSMs
\cite{Chen05,Cotter05,Baron05,Marco05a,Tropp06,Eldar08,Duarte09}.
Most of them \cite{Chen05,Cotter05,Baron05,Marco05a,Chen06,Tropp06}
focus on a JSM
in which the column vectors $\vecg\alpha^{(j)}$
simply share their support $\mathcal{T}$.
This JSM is the special case of our row-block sparsity model with
$\mathcal{N}=\mathcal{N}_1$ described in Section \ref{subsec:formulation}.
Efficient algorithms for the MMV problem with this JSM
have been designed 
as the extensions of greedy algorithms such as
matching pursuit (MP) and
orthogonal matching pursuit (OMP)
\cite{Pati93,Davis97,Tropp04,Donoho06robust,Tropp07}.
OMP is an efficient algorithm that can recover
a $m$-sparse vector from
a $\mathcal{O}(m\log N)$-dimensional vector \cite{Tropp07}.
It iteratively selects the basis (column of $\mtr A$)
with the largest contribution to the current residual
to reduce greedily the representation error at each iteration.
The existing MP- and OMP-based algorithms for the MMV problem
can be directly used for our problem with the row-block sparsity model
only when $\mathcal{N}=\mathcal{N}_1$.

Eldar and Mishali \cite{Eldar08}
introduced the block sparsity model and block RIP condition
applicable to MMV problems including ours.
The uniqueness was guaranteed in \ref{subsec:formulation}.
By $l^1$ convex relaxation,
we can cast the vectorized version
in (\ref{eq:sparse decomposition by minimization vectorized}) as
\begin{eqnarray}
&& \min_{\mtr A} ||\vectorize(\mtr A^\top)||_{1,\mathcal{N}'}
\quad\mbox{subject to}\quad\nonumber\\
&&\qquad\qquad\vectorize(\mtr Q^\top)=(\mtr S\otimes\mtr I_n)\vectorize(\mtr A^\top).
\label{eq:sparse decomposition by l1 minimization vectorized}
\end{eqnarray}
Here, we redefine $f_{\mathcal{N}}$ as a map from
$(\mathbb{R}^{N\times n},l^F)$ to $(\mathbb{R}_{+}^C,l^1)$
in the same form as (\ref{eq:map to l0 classification space}),
and define
\footnote{The norm $||\cdot||_{2,\mathcal{I}}$ defined in \cite{Eldar08} is the same as our $||\cdot||_{1,\mathcal{N}}$, and it is actually the $l^1$ norm through $f_{\mathcal{N}}$ as we defined.}
\begin{equation}
 ||\mtr A||_{1,\mathcal{N}}\defas||f_{\mathcal{N}}(\mtr X)||_1.
\end{equation}
According to \cite{Eldar08},
this $l^1$ minimization problem is a second order cone problem (SOCP).

\subsection{Sub-optimal algorithm}

\begin{algorithm}[H]
\caption{Sparse subspace decomposition (SSD-ROMP)}
\label{alg:SSD-ROMP}
\begin{algorithmic}[1]
\REQUIRE
$\mtr Q\in\mathbb{R}^{d\times n}$: matrix of $n$ queries
as (\ref{eq:query matrix}),\quad
$\mtr S\in\mathbb{R}^{d\times N}$: concatenated matrix of training datasets as (\ref{eq:concatenated training datasets}),\quad
$\mathcal{N}$: list of row block sizes,\quad
$M_0$: sparsity level;
\ENSURE
$\mtr A$: row-block sparse matrix as (\ref{eq:row stacked form}),
$\mathcal{I}$: set of indices of nonzero blocks;
\STATE
let the index set $\mathcal{I}\defas\emptyset$ and residual $\mtr R\defas\mtr Q$;
\REPEAT
\STATE
$\mtr U\defas\mtr S^\top\mtr R$;\label{step:u} 
\STATE
$\vecg\gamma\defas f_{\mathcal{N}}(\mtr U)$; 
\STATE
let $\mathcal{J}$ be a set of indices of
the $M_0$ biggest components of $\vecg\gamma$,
or all of its nonzero components, whichever set is smaller; 
\STATE
sort $\mathcal{J}$ in descending order of the components $\vecg\gamma$; 
\STATE
among all subsets $\mathcal{J}_0\subset\mathcal{J}$ 
such that $\gamma_i\leq 2\gamma_j$ for all $i<j\in\mathcal{J}_0$,
choose $\mathcal{J}_0$ with the maximal energy
$||\gamma|_{\mathcal{J}_0}||_2^2\defas \displaystyle\sum_{k\in\mathcal{J}_0}\gamma_k^2$; 
\STATE
$\mathcal{I}\defas\mathcal{I}\cup\mathcal{J}_0$;
\FOR{each $j$}
\STATE
 $\vecg\alpha^{(j)}\defas\displaystyle\arg\min_{\vecg\alpha}||\vec q^{(j)}-\sum_{k\in\mathcal{I}}\mtr S_k\vecg\alpha||_2$;\label{step:lsq} 
\ENDFOR
\STATE
$\mtr R\defas\mtr Q-\displaystyle\sum_{k\in\mathcal{I}}\mtr S_k\mtr A_k$; 
\UNTIL
$||\mtr R||_F=0$ or $\card\mathcal{I}\geq 2M_0$.
\end{algorithmic}
\end{algorithm}

\begin{figure*}[htb]
\setlength{\unitlength}{1cm}
\begin{center}
\begin{picture}(16,10.7)
\put(-0.8,7.7){\includegraphics[height=3cm]{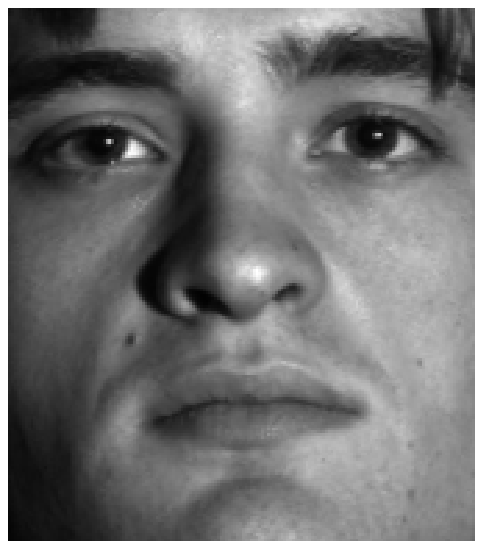}}
\put(-0.8,4.3){\includegraphics[height=3cm]{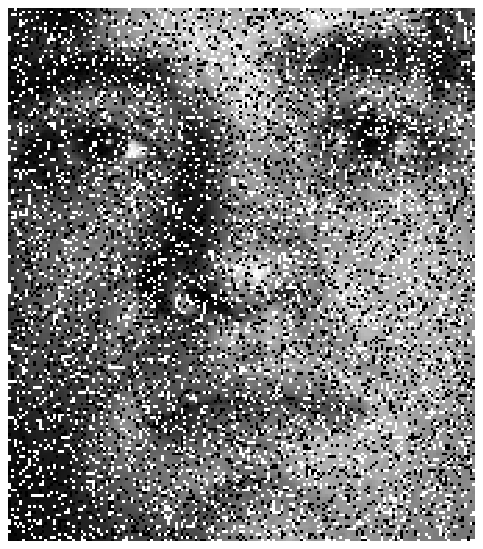}}
\put(-0.8,0.9){\includegraphics[height=3cm]{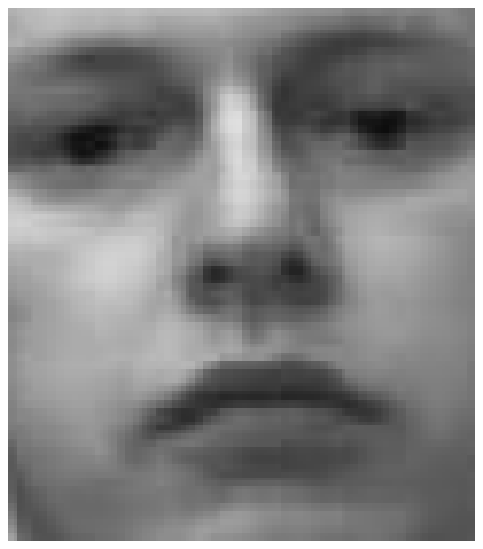}}
\put(2,7.7){\includegraphics[height=3cm]{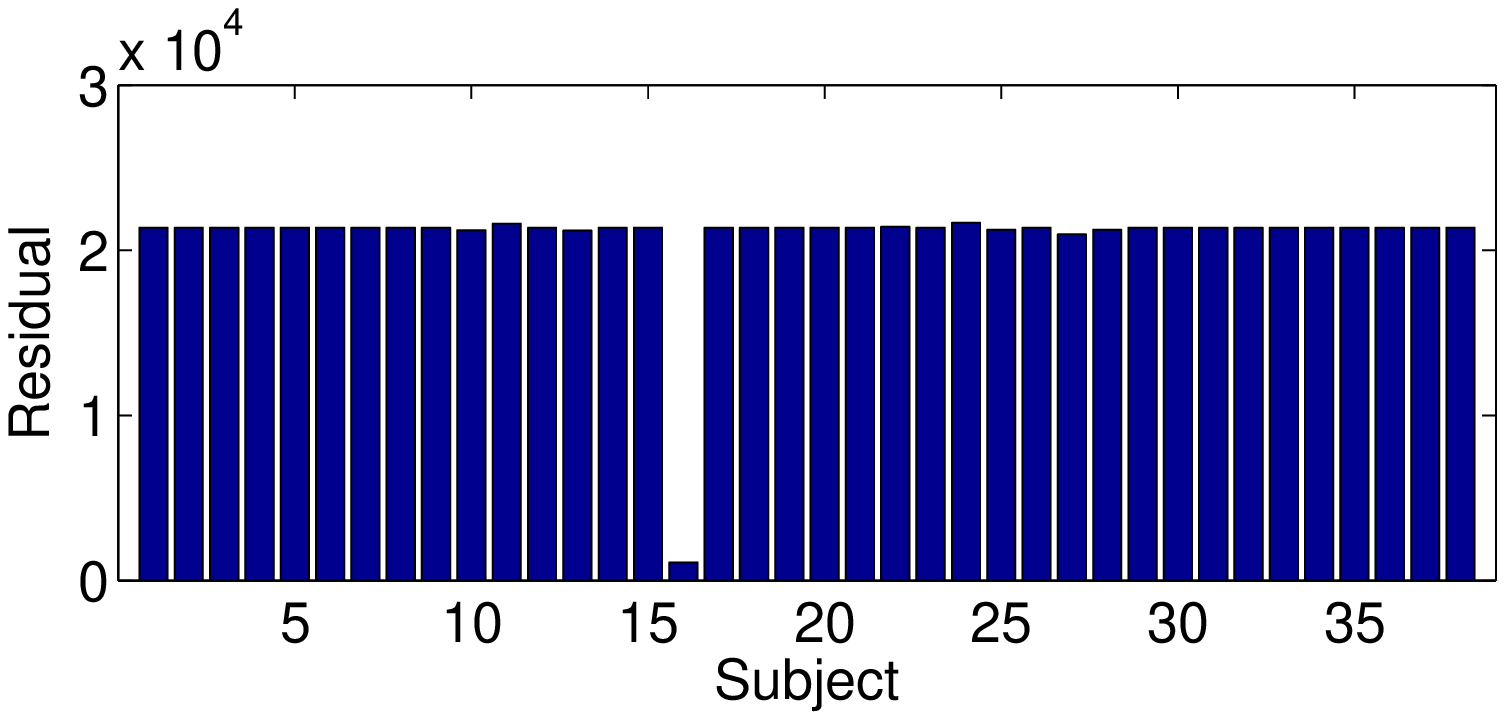}}
\put(9,7.7){\includegraphics[height=3cm]{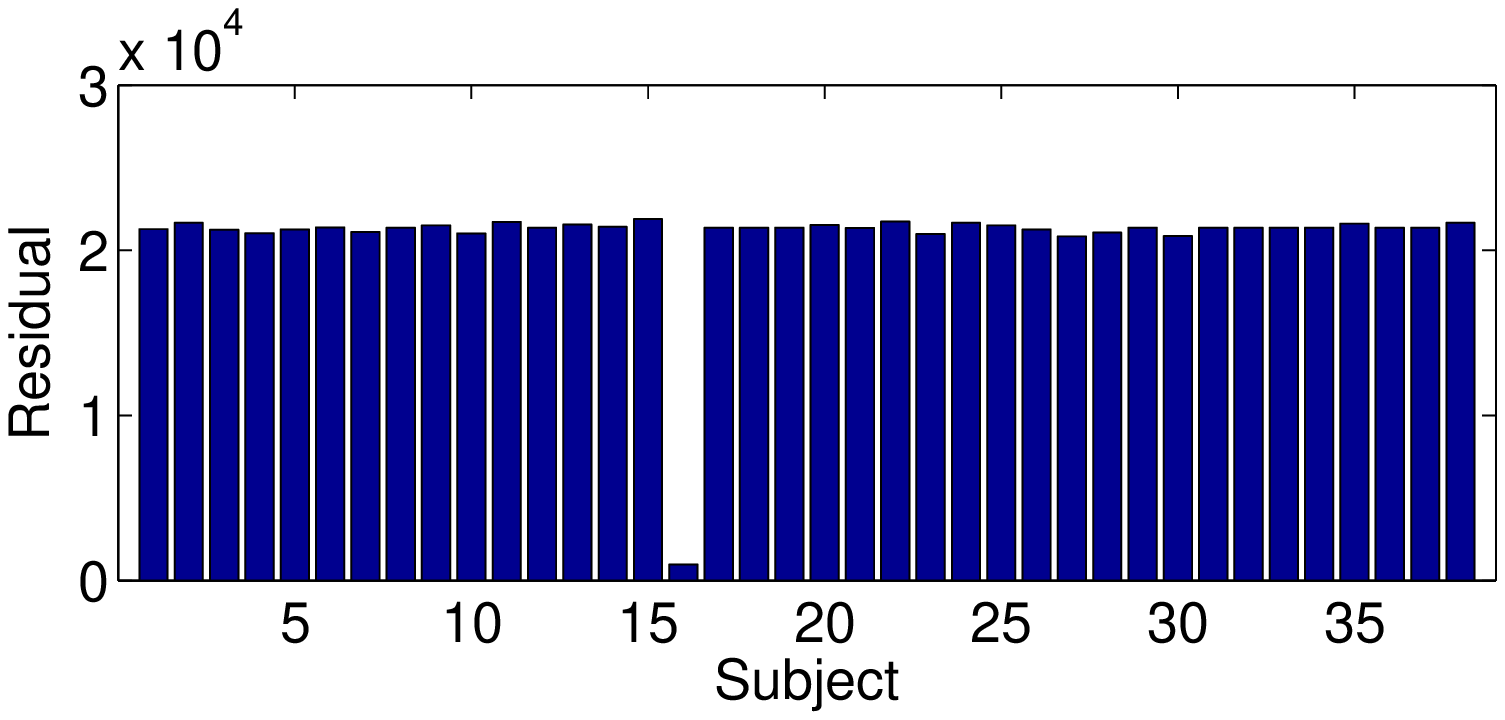}}
\put(2,4.3){\includegraphics[height=3cm]{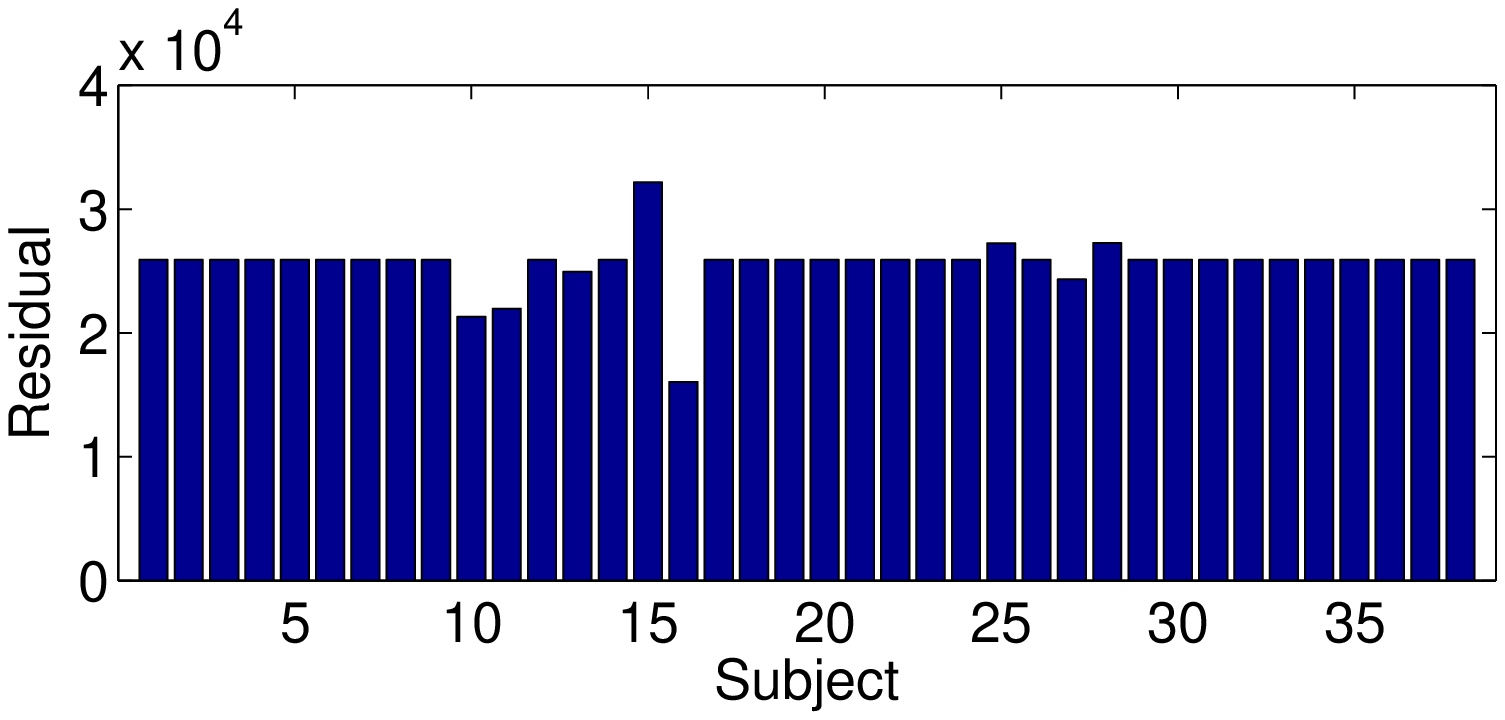}}
\put(9,4.3){\includegraphics[height=3cm]{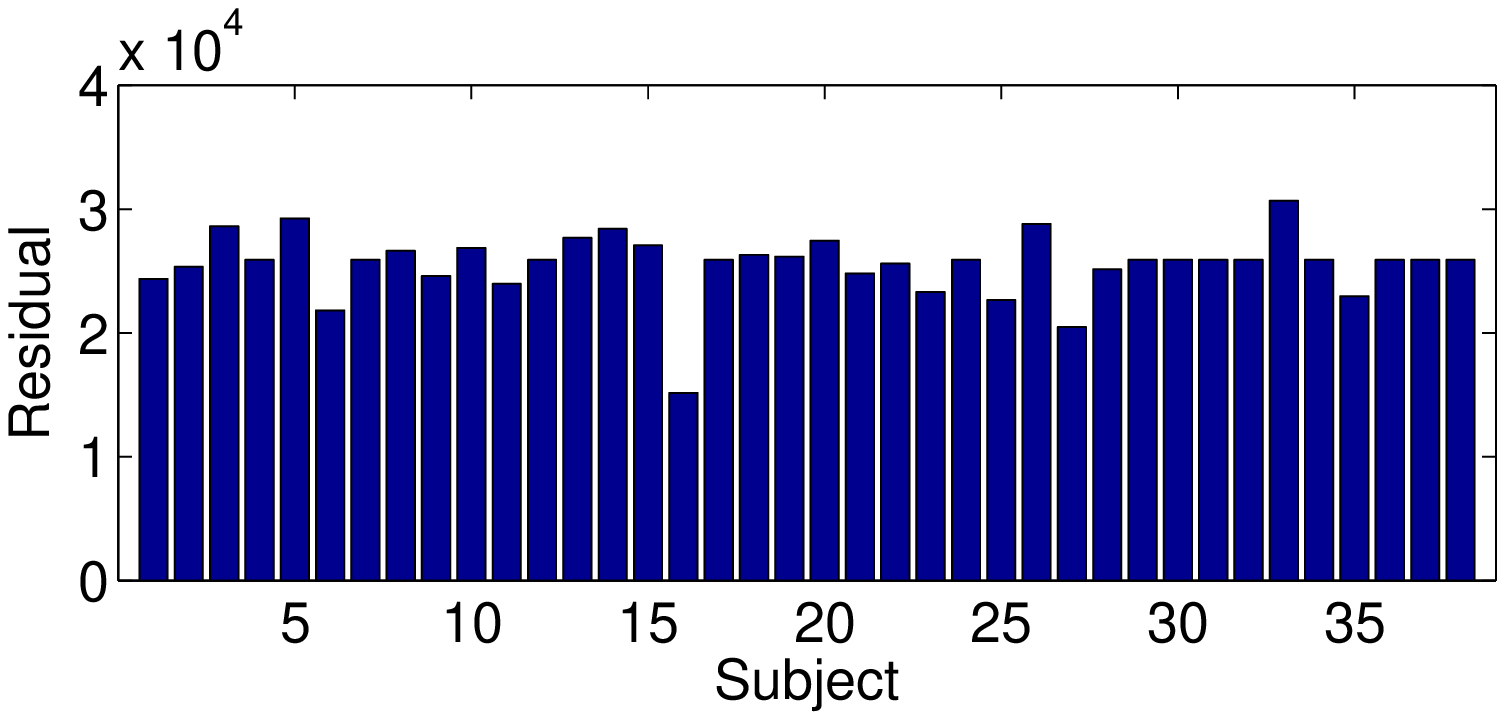}}
\put(2,0.9){\includegraphics[height=3cm]{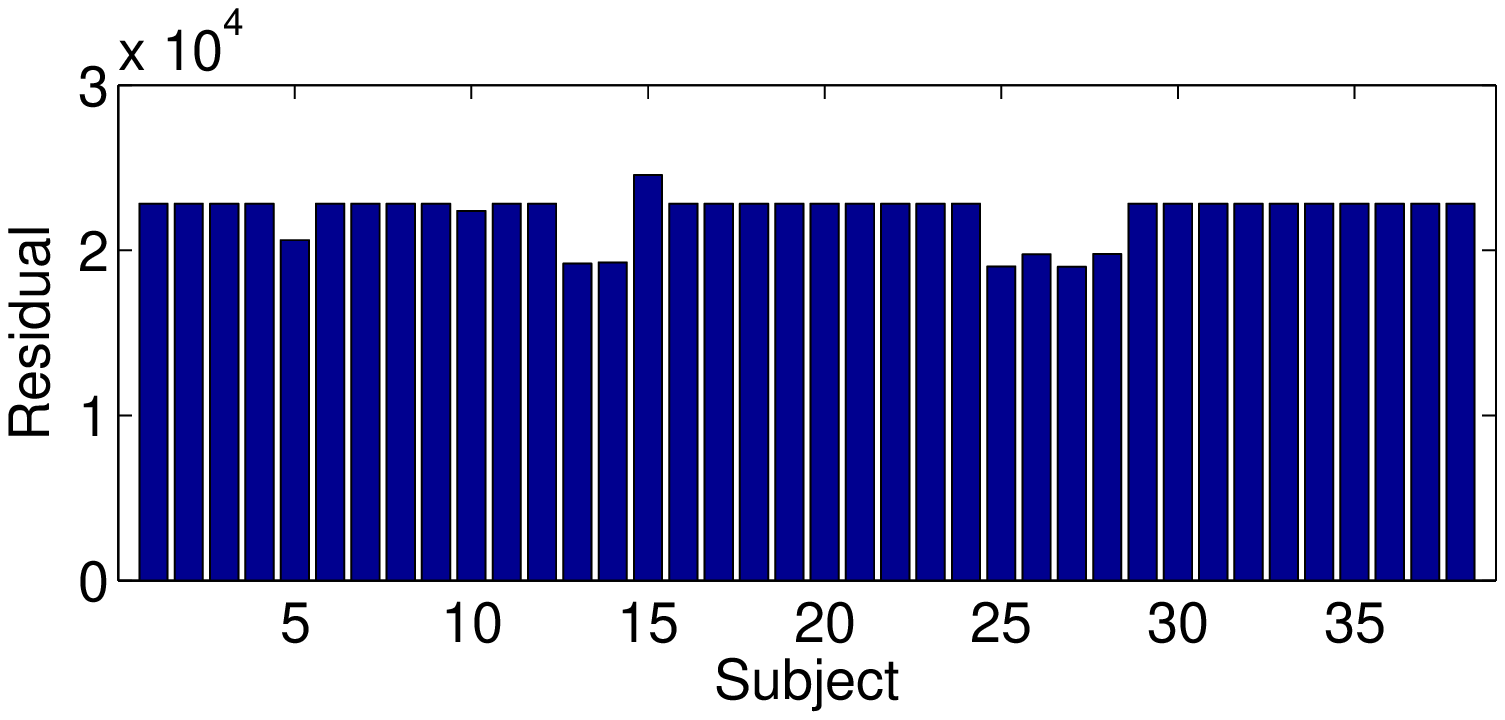}}
\put(9,0.9){\includegraphics[height=3cm]{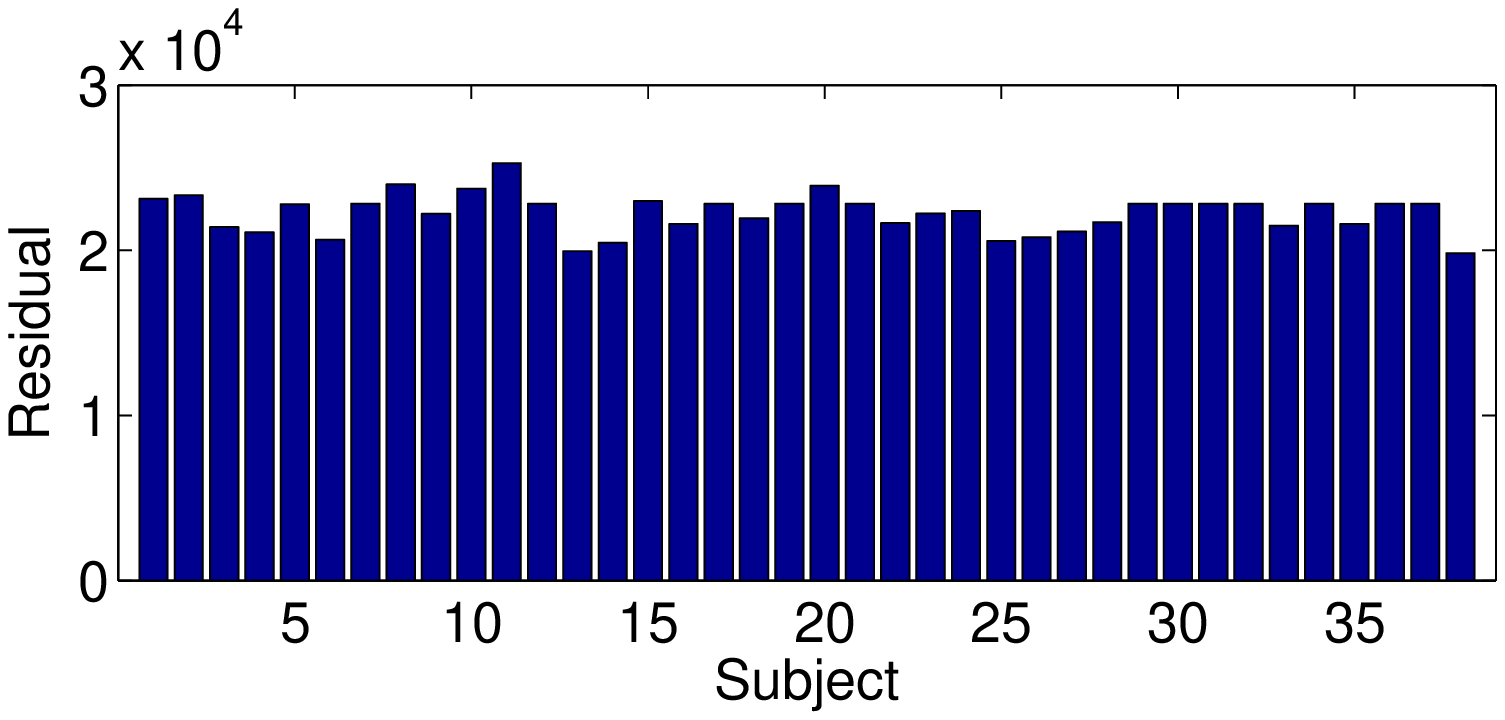}}
\put(0.8,7.7){(a)}
\put(0.8,4.3){(b)}
\put(0.8,0.9){(c)}
\put(5.5,0.2){SSM}
\put(12.5,0.2){SRC}
\end{picture}
\caption{Examples of one-to-one classification.
The first, second, and third columns respectively show the query images,
residuals of the representations by SSM with respect to 38 sujects,
and those by SRC.
(a) A valid query image of subject \#16.
(b) The same query image as (a) with 30\% pixels corrupted
by salt and pepper noise.
(c) An invalid image from unlearned face database.}
\label{fig:YaleB redisuals salt papper}
\end{center}
\end{figure*}

We present a practical greedy algorithm of
the block sparse decomposition.
Although
there are optimization packages
that solve the SOCP in polynomial time,
we prefer a simple and efficient algorithm
of the sparse recovery
like the MP and OMP.
As compared with the signal recovery in compressed sensing,
approximate solutions may be enough
for the classification purpose.
Since the sparsity level is at most
$\mathcal{O}(n)$ for $n$ queries,
we want the decomposition algorithm to work efficiently
in the case of extreme sparseness.

We adopt the regularized OMP (ROMP) \cite{Needell09}
because it can stably provide
approximate solution from noisy queries.
We modify the ROMP to seek for the nonzero row blocks of
the solution as shown in Algorithm \ref{alg:SSD-ROMP}.
This algorithm
selects multiple row-blocks of $\mtr S^\top\mtr S\mtr A$
that have comparable magnitudes measured by $f_\mathcal{N}$
at each iteration.
Note that the algorithm requires the additional parameter
$M_0=\mathcal{O}(M)$ although the solution is insensitive to this parameter.

Intensive computations are the matrix multiplication at Step \ref{step:u}
and the least squares problem at Step \ref{step:lsq},
which cost $\mathcal{O}(nNd)$ and $\mathcal{O}(nM_0^2d)$ time, respectively.
The cost of least squares problem can be reduced to
$\mathcal{O}(nM_0d)$ by the conjugate gradient (CG) method
as suggested in \cite{Needell09}.
The total running time of Algorithm \ref{alg:SSD-ROMP}
is $\mathcal{O}(nM_0^2Nd)$ or $\mathcal{O}(nM_0Nd)$ using CG.

\section{Experiments}
\label{sec:experiments}

We demonstrate our sparse subspace method (SSM)
described in Algorithm \ref{alg:SSM}.
We perform face recognition experiments using a cropped version
of the Extended Yale Face Database B \cite{GeBeKr01,KCLee05}.
The database consists of 2,414 face images of 38 individuals.
We randomly select half of the images of each subject
for the training dataset ($n_k\approx 32$, $k=1,\dots,38$),
and the other half for queries.
Each image is expressed as a $d=192\times 168=32,\!256$ dimensional vector
storing the grayscale values.

\paragraph{One-to-one classification}

Figure \ref{fig:YaleB redisuals salt papper}
shows examples of one-to-one classification.
The SSM tries to answer a class label for a single query.
We reduced the dimensionality to $\hat d=1,\!024$
by the Gaussian random projection
at Step \ref{step:dimensionality reduction} in Algorithm \ref{alg:SSM}.
We set the block sizes $\mathcal{N}=\{n_1,\dots,n_{38}\}$
and the sparsity level $M_0=4$ in Algorithm \ref{alg:SSD-ROMP}.
Since SSM behaves as the SRC \cite{Wright08}
when $\mathcal{N}=\mathcal{N}_1$,
we also executed the SRC implemented with ROMP.
The SSM and SRC, including the random projection,
run in less than 0.2 seconds on a moderate workstation.

For the valid query image of subject \#16
as Fig. \ref{fig:YaleB redisuals salt papper}(a),
we see that only the residual $r_{16}$ is significantly small.
The SSM and SRC stably detect $r_{16}$ as the smallest
even if the query is contaminated with noise
as shown in Fig \ref{fig:YaleB redisuals salt papper}(b)
before the dimensionality reduction.
We also observe in Fig \ref{fig:YaleB redisuals salt papper}(c)
that none of the residuals can be significantly small
for the invalid query (taken from the UMIST face database \cite{UMIST}).
In all cases, the residuals tend to be left undisturbed in SSM
although the classification results are the same as SRC.
This indicates that irrelevant class subspaces are ruled out
by the block sparse model.

\paragraph{$n$-to-one classification}

\begin{figure}[htb]
\setlength{\unitlength}{1cm}
\begin{center}
\begin{picture}(8,6)
\put(0,0){\includegraphics[height=6cm]{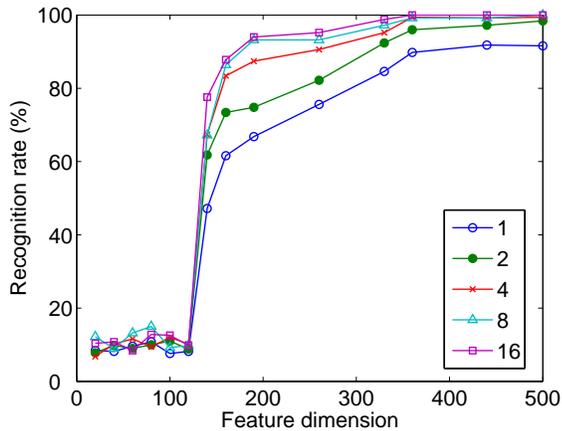}}
\end{picture}
\caption{Recognition rates of $n$-to-one classifier
on Extended Yale B database, with respect to feature dimension.}
\label{fig:YaleB recognition rate vs dim}
\end{center}
\end{figure}

For different numbers $n$ of queries,
we evaluated the recognition rate of $n$-to-one classifier
with respect to reduced feature dimension $\hat d$
by Gaussian random projection.
For $\hat d>120$,
the recognition rate increases with $n$ and $\hat d$
as shown in Fig. \ref{fig:YaleB recognition rate vs dim}.
The rate is enhanced to more than 99\% at $\hat d>350$
with $n\geq 4$ queries.
The perfect classification is achieved at $\hat d>400$
with $n\geq 8$ queries.
The $n$-to-one classifier provides
better performance than the one-to-one classifier
applied to each query,
because the $n$-to-one classifier
takes advantage of the joint sparsity.
However,
the SSM did not improve the recognition rate
at low dimensions $\hat d<120$.
We should cope with this matter in the future work.

\paragraph{$n$-to-ones classification}

\begin{figure}[htb]
\setlength{\unitlength}{1cm}
\begin{center}
\begin{picture}(7.8,8.8)
\put(-0.6,0){\includegraphics[height=9.5cm]{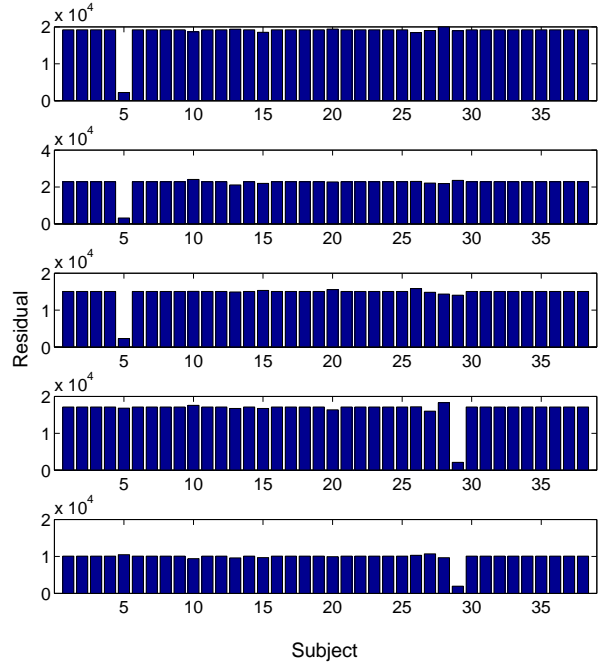}}
\end{picture}
\caption{An example of $n$-to-ones classification.
Residuals $r_k^{(1)},\dots,r_k^{(5)}$
are shown from top to bottom.
Each of $n=5$ queries is  classified into one of two classes $k=5$ and $29$.}
\label{fig:YaleB multiple face classification}
\end{center}
\end{figure}

We also performed the $n$-to-ones classification.
Figure \ref{fig:YaleB multiple face classification}
shows an example using the Extended Yale Face Database B.
We gave the classifier five query images,
three of which are taken from subject \# 5
and two from \# 29.
These five queries are
classified into their respective classes
indicated by the significantly small residuals.

\section{Concluding remarks}
\label{sec:conclusion}

We have developed the sparse subspace method (SSM),
which enables us to classify multiple queries
into their respective classes, simultaneously.
The SSM is based on the sparse decomposition of the query subspace.
The query subspace is represented
only by the relevant class subspaces.
Since this sparse decomposition can be cast as
the MMV problem with a row-block joint sparsity model,
the uniqueness, robustness and recovery of the solution are guaranteed
under the block RIP condition.
We realized the block sparse decomposition
by modifying the greedy algorithm ROMP.
We experimentally showed that the classification of multiple queries
improves the recognition rate on a face database.
The joint sparsity model
and the decomposition algorithm should be improved further.
More detailed performance evaluation also remains in the future work.

{\small
\bibliographystyle{ieeetr}
\bibliography{mybib}
}

\end{document}